\documentclass[conference]{IEEEtran}
\IEEEoverridecommandlockouts
\IEEEaftertitletext{\vspace{-1.5\baselineskip}}
\ifCLASSOPTIONcompsoc
  \usepackage[nocompress]{cite}
\else
  \usepackage{cite}
\fi
\usepackage{amsmath,amssymb,amsfonts}
\usepackage{algorithmic}
\usepackage{graphicx}
\usepackage{textcomp}
\usepackage{xcolor}
\usepackage{url}
\usepackage{arydshln}
\usepackage{dsfont}
\usepackage{pifont}
\usepackage{graphicx}
\usepackage{multirow}

\def\BibTeX{{\rm B\kern-.05em{\sc i\kern-.025em b}\kern-.08em
    T\kern-.1667em\lower.7ex\hbox{E}\kern-.125emX}}

\usepackage[normalem]{ulem}
\newcommand{\rev}[1]{#1}

\newcommand{\revtwo}[1]{#1}

\usepackage{tikz}
\usepackage{textcomp}
\usepackage[doipre={DOI:~}]{uri}
\usepackage{lipsum}
\newcommand\copyrighttext{%
  \footnotesize \textcopyright 2024 IEEE. Personal use of this material is permitted.  Permission from IEEE must be obtained for all other uses, in any current or future media, including reprinting/republishing this material for advertising or promotional purposes, creating new collective works, for resale or redistribution to servers or lists, or reuse of any copyrighted component of this work in other works.
  }
\newcommand{\copyrightnotice}{%
\begin{tikzpicture}[remember picture,overlay]
\node[anchor=south,yshift=10pt] at (current page.south) {\fbox{\parbox{\dimexpr\textwidth-\fboxsep-\fboxrule\relax}{\copyrighttext}}};
\end{tikzpicture}%
}

\definecolor{somegray}{rgb}{0.5, 0.5, 0.5}
\newcommand{\darkgrayed}[1]{\textcolor{somegray}{#1}}
\makeatletter
\newcommand*\titleheader[1]{\gdef\@titleheader{#1}}
\AtBeginDocument{%
  \let\st@red@title\@title
  \def\@title{%
    \vskip-2.0em
    \bgroup\normalfont\large\centering\@titleheader\par\egroup
    \vskip0.0em\st@red@title}
}

\makeatother

\titleheader{\darkgrayed{This paper has been accepted for publication in the  Sixth IEEE International Conference on Image Processing Applications and Systems\\\copyright 2024 IEEE.}}

\title{Building Damage Assessment in Conflict Zones: A Deep Learning Approach Using Geospatial Sub-Meter Resolution Data\vspace{-0.6cm}
\thanks{This work has received funding from the ESA ``Space in Response to Humanitarian Crises" program (ESA Contract 4000142151/23/NL/EG/an).\\ We also thank Nil Pedro Angli of ESA, Bruno Aster and Elena Lorusso of Gisky s.r.l., Filippo Dacarro and Luca Grottoli of Fondazione Eucentre.}
}
\begin{document}
\bstctlcite{IEEEexample:BSTcontrol}

\author{
\IEEEauthorblockN{Matteo Risso\IEEEauthorrefmark{1}, Alessia Goffi\IEEEauthorrefmark{2}, Beatrice Alessandra Motetti\IEEEauthorrefmark{1}, Alessio Burrello\IEEEauthorrefmark{1}, Jean Baptiste Bove\IEEEauthorrefmark{3},\\Enrico Macii\IEEEauthorrefmark{1}, Massimo Poncino\IEEEauthorrefmark{1}, Daniele Jahier Pagliari\IEEEauthorrefmark{1}, Giuseppe Maffeis\IEEEauthorrefmark{2}}
\IEEEauthorblockA{\IEEEauthorrefmark{1}Politecnico di Torino, Italy, \textit{Email: firstname.firstsurmame@polito.it}\\
\IEEEauthorrefmark{2}TerrAria s.r.l., Italy. \textit{Email: firstlettername.surname@terraria.com} \IEEEauthorrefmark{3}Croce Rossa Italiana, Italy. \textit{Email: jeanbaptiste.bove@cri.it}}
}










\maketitle
\copyrightnotice

\begin{abstract}
Very High Resolution (VHR) geospatial image analysis is crucial for humanitarian assistance in both natural and anthropogenic crises, as it allows to rapidly identify the most critical areas that need support. Nonetheless, manually inspecting large areas is time-consuming and requires domain expertise.
\rev{Thanks to their accuracy, generalization capabilities, and highly parallelizable workload, Deep Neural Networks (DNNs) provide an excellent way to automate this task. Nevertheless, there is a scarcity of VHR data pertaining to conflict situations, and consequently, of studies on the effectiveness of DNNs in those scenarios.
Motivated by this, our work extensively studies the applicability of a collection of state-of-the-art Convolutional Neural Networks (CNNs) originally developed for \textit{natural disasters} damage assessment in a war scenario.
To this end, we build an annotated dataset with pre- and post-conflict images of the Ukrainian city of Mariupol. 
We then explore the transferability of the CNN models in both zero-shot and learning scenarios, demonstrating their potential and limitations.
To the best of our knowledge, this is the first study to use sub-meter resolution imagery to assess building damage in combat zones.}
\end{abstract}

\section{Introduction}~\label{sec:intro}
%
In recent years, the usage of geospatial images has been studied as a key element to help and speed-up humanitarian assistance in crisis situations~\cite{xBD, kyiv_1, kyiv_2, mosul}. In particular, the ability to quickly and accurately assess damage to buildings and infrastructures is crucial for effective response and recovery. Traditional methods of damage assessment, which rely heavily on manual inspection and reporting, are often slow and resource-intensive~\cite{survey_pixel_obj}. 
Automated approaches using Deep Neural Networks (DNNs) offer a promising alternative, enabling rapid and scalable analysis of large volumes of geospatial data. 
\rev{When compared to other Computer Vision (CV) algorithms, DNNs have demonstrated exceptional performance and robustness to variations in image quality, lighting conditions and angles, thanks to their ability to learn and extract relevant features from raw images~\cite{cv_vs_cnn}. Additionally, DNN workloads are highly scalable and parallelizable, enabling the rapid processing of large volumes of data.}
\looseness=-1
For these reasons, they are considered state-of-the-art (SotA) for automated image analysis tasks such as segmentation, classification, and object detection~\cite{unet_original, SE}. Building and infrastructure damage may be quickly identified and measured by utilizing the capabilities of these models~\cite{wheeler2020deep, chen2022dual, shen2021bdanet}. 

\looseness=-1
Despite the potential of DNNs, a significant challenge remains: the availability of high-quality, annotated data. While extensive data sources of Very High Resolution (VHR) satellite imagery for natural disasters~\cite{xBD} exist, data specific to war scenarios is comparatively scarce. This gap hinders the development of DNN models tailored for conflict-related damage assessment. 
%
\rev{With this limitation in mind, our research aims to narrow the gap through extensive experimentation aimed at verifying whether DNN models originally designed for natural damage detection, can be effectively employed also for war-induced damages. To test this hypothesis, we collect an annotated dataset focused on the Mariupol area in Ukraine, comprising pre- and post-conflict images with sub-meter resolution.}
On this dataset, we assess a collection of SotA Convolutional Neural Network (CNN) models~\cite{xview_winner}, originally developed for damage assessment in natural disaster scenarios~\cite{xBD}. We evaluate their ability to transfer learned features to the context of war-induced damage, exploring both \textit{zero-shot} scenarios, where the models are applied to the new dataset without additional training on, and \textit{learning scenarios}, where the models are fine-tuned on the Mariupol data. This approach allows us to assess the transferability and adaptability of these models, providing insights into their performance and identifying potential areas for improvement.

\rev{
The main contributions of this work are summarized below:
\begin{itemize}
    \item We extensively test the performance of four SotA CNN models, originally proposed for building damage assessment after a natural disaster, in a war scenario, testing and comparing zero-shot and transfer learning setups.
    \item We propose a custom data augmentation pipeline to improve the accuracy of fine-tuned models on a relatively small dataset from the Ukranian city of Mariupol.
    \item We propose an ablation study highlighting the importance of data augmentation and of the dilation filter pre-processing.
    \item Overall, our models achieve up to 69\% and 59\% F1 score and up to 86\% and 79\% balanced accuracy score on 2- and 3-class versions of the damage assessment problem.
\end{itemize}
}
%

\vspace{-0.2cm}
\section{Background \& Related Work}~\label{sec:background_related}
%
%
\vspace{-0.5cm}
\subsection{Data Sources}
Humanitarian Assistance and Disaster Response (HADR) relies heavily on the availability of VHR (i.e., with $<2$m resolution) geospatial data sources~\cite{xBD}.
Mainly, satellite imagery is used, allowing data collection over large areas. 
An alternative is represented by images collected from Unmanned Aerial Vehicles (UAVs) which can achieve higher resolution despite a smaller area coverage~\cite{uav_ref}.
In this work, we concentrate on satellite imagery since drone use in conflict zones may be limited due to security and airspace restrictions.

Commercial satellite data providers employ proprietary satellites such as WorldView-3 and 4 by Maxar or the Pléiades sensors by Airbus Defense.
Public data providers include government agencies such as NASA through the Landsat program or the ESA with the Sentinel-2 satellite. In both cases, despite being free and openly available, the provided resolution is lower when compared to commercial solutions~\cite{satellite_comp}.

\looseness=-1
xBD~\cite{xBD} represents the only large-scale dataset containing pre- and post-disaster geospatial VHR images, provided freely by Maxar, with more than 800k annotated buildings. The dataset was developed in the context of the xView2 challenge promoted by the Defense Innovation Unit of the US Department of Defense. It includes RGB data with a resolution of 0.8 meters, relative to 19 different disasters from all over the world, thus showing great variability in terms of location and disaster type. For each pair of pre- and post-disaster images, the footprint mask of all buildings is provided, coupled with a multi-damage scale including no damage, minor damage, major damage, and destroyed.
To the best of our knowledge, no open and freely available dataset exists with annotated damage in war scenarios.
\subsection{State-of-the-Art}
Building damage assessment and more in general Change Detection (CD) utilizing VHR geospatial images has been the subject of several research works.
The two main families of approaches include pixel-based and object-based algorithms~\cite{survey_pixel_obj}. Historically, the first approaches were pixel-based and were limited mainly to low- and medium-resolution imagery. \rev{Pixel-based techniques include texture-based analysis, where the amount of change is assessed comparing each pixel with image-wise statistics, using deterministic thresholds~\cite{texture_analysis}. Fuzzy CD~\cite{fuzzy} uses a similar pipeline coupled with fuzzy reasoning to express thresholds in terms of probabilities.}
The application of traditional pixel-based techniques in the high-resolution regime is hindered by the strong dependence of such methods on reflectance variability and different acquisition characteristics~\cite{survey_pixel_obj}.
\looseness=-1
Conversely, object-based CD involves segmenting an image into meaningful objects and classifying them based on their texture, shape, spatial arrangement, and spectral properties~\cite{survey_pixel_obj}. 

With Deep Learning (DL), the popularity of pixel-based techniques has been revamped. Indeed, the current SotA is mostly made of pixel-based methods leveraging DNNs.
\cite{unet++} proposes a CNN to perform CD over multiple years in an urban scenario. 
Similarly, BLDNET~\cite{ismail2022bldnet} performs CD employing a Graph Convolutional Network augmented with urban knowledge.
\cite{wheeler2020deep} and \cite{chen2022dual} try to solve the damage assessment problem using respectively CNN and transformer-based architectures and test the proposed approaches on the xBD dataset.
On the same dataset, BDANet~\cite{shen2021bdanet} proposes a hybrid CNN-transformer architecture.
As said, most of these works consider generic CD scenarios, while a limited subset addresses the damage assessment problem with experiments solely on the xBD dataset.
\rev{Conversely, to the best of our knowledge, no DNN methods have been specifically proposed for war-related scenarios, making our approach the first of its kind.}
The only approaches targeting such scenarios are based on shallow algorithms.
\cite{mosul} employs Sentinel-1 and -2 data to perform a CD study in the war conflict area of Mosul using a Support Vector Machine (SVM) classifier and texture analysis.
\looseness=-1
\cite{kyiv_1} and \cite{kyiv_2} consider the city of Kyiv as case-study. In both cases, Sentinel data are used to assess damage. In \cite{kyiv_1}, Sentinel-1 data are used to perform an intensity analysis while texture analysis is performed on Sentinel-2 data. \cite{kyiv_2} extracts features from windows of the original images as time series and employs the Pruned Exact Linear Time algorithm to identify change points.
%

\section{Methods}~\label{sec:methods}
\vspace{-0.6cm}
\subsection{Data sources and Dataset Creation}~\label{subsec:dataset}
The objective of this work is to test whether DL models designed for generic damage assessment in natural disasters can transfer well to war-induced damage scenarios.
To this end, we collect a new dataset for building damage assessment in the context of the Russo-Ukrainian war. The data has been acquired from Airbus Defense using their proprietary Pléiades sensor. The starting points are two VHR images of the Mariupol area,
%
one captured on July 11th, 2021, prior to the conflict, and the other on August 6th, 2022, subsequent to the onset of hostilities.
\revtwo{The location has been selected based on multiple reasons: the availability of archival VHR images, evidence of widespread destruction, and the interest for humanitarian assistance associations.}
The city of Mariupol, in the Donetsk Oblast, has an area of over 150 square kilometers with a population of 446,103 people. The analysis focused on the most affected area, of approximately 16 square kilometers.
The Pléiades sensor offers a ground resolution of 0.5 meters, ensuring that even small-scale damage can be accurately assessed. The spectral bands included in the dataset are Blue (430-550nm), Green (480-610nm), Red (600-720nm), and Near-infrared (750-950nm). Following the SotA~\cite{xBD} we employ only the RGB bands. Panchromatic Multispectral Sharpening (PMS) is applied to produce high-resolution color images by combining the lower-resolution multispectral bands with the higher-resolution panchromatic band.
\rev{Then, the images are orthorectified with a low-resolution digital elevation model (indicatively at 30m)} to correct geometric distortions caused by sensor tilt and terrain relief.
%
\looseness=-1
Finally, the images are processed to conform to the Universal Transverse Mercator (UTM) projection, referenced to the World Geodetic System 1984 (WGS84). The processed images are stored in TIFF format.

\begin{figure}
  \centering
  \includegraphics[width=0.95\columnwidth]{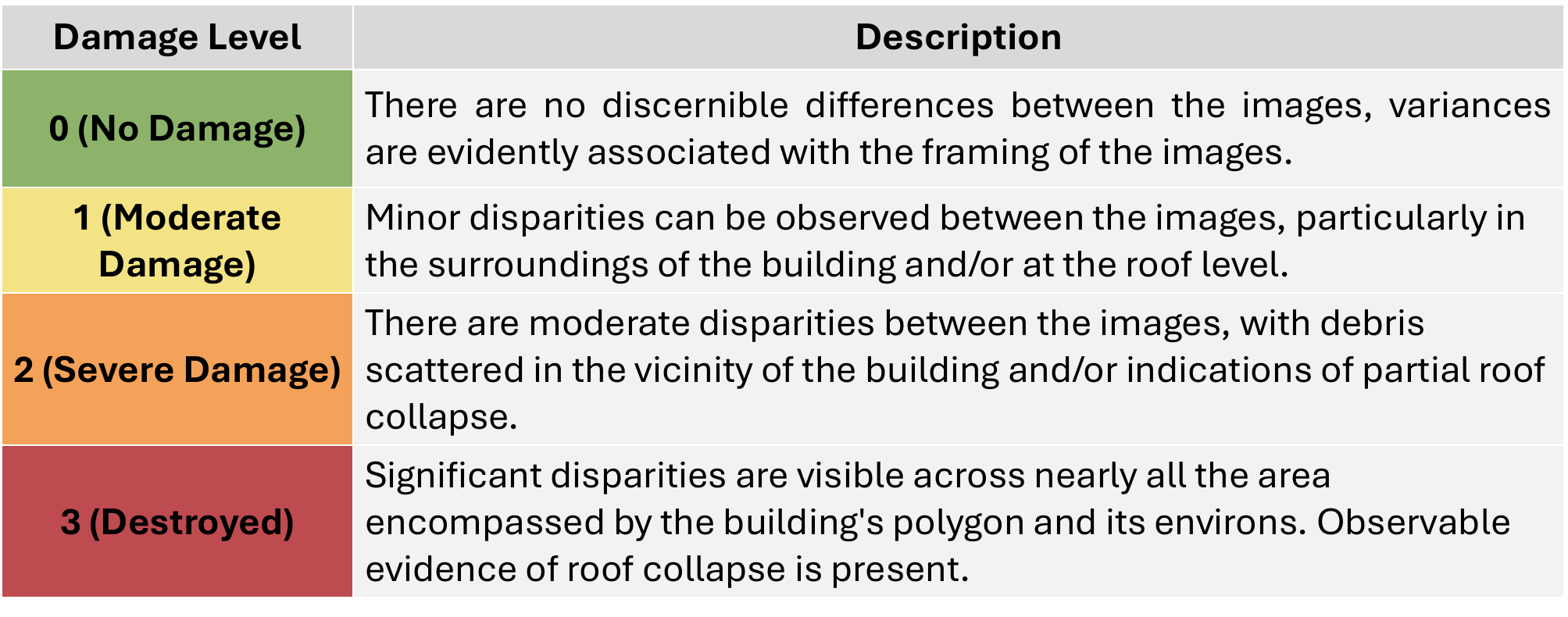}
  \vspace{-0.4cm}
  \caption{Damage scale description.}
  \label{fig:damage_scale}
  \vspace{-0.3cm}
\end{figure}

The two VHR images, coupled with building footprints provided by the OpenStreetMap (OSM) platform~\cite{OSM}, are used to systematically compare and assess the damage undergone by each building.
Data labelling has been performed manually by civil engineering domain experts following the damage levels criteria of Fig.~\ref{fig:damage_scale}.
Moreover, the classification database has been enhanced with the analysis made by the United Nations Institute for Training and Research (UNITAR)\footnote{\url{https://www.unitar.org/}} on the same area, based on imagery collected on March 14th, May 7th, 8th, 12th, 2022 and June 21st, 2021 available on their website.
\rev{The labeling of the buildings was performed in 2 steps, firstly using the information available from UNITAR and secondly classifying the missing buildings by photo-interpretation by the domain experts. To assign the degree of damage derived from UNITAR (available in terms of georeferenced points) to the buildings in the OSM database, a buffer of 7.5m was applied around the ground footprint of the OSM building, and a spatial correspondence was created with the points falling within this envelope.  If a single point was assigned to several buildings, the UNITAR point was associated with the building with the closest distance from the barycentre of the polygon. In case of multiple points contained in the same building, the most severe degree of damage was selected. The buffer of 7.5m was assessed as the best compromise in terms of point allocation and reduction of the above-mentioned cases. The second step was then carried out instructing experts to be as homogeneous as possible with the UNITAR classification in terms of allocation and assessment of the degree of damage to buildings.}
\begin{figure}
  \centering
  \includegraphics[width=0.8\columnwidth]{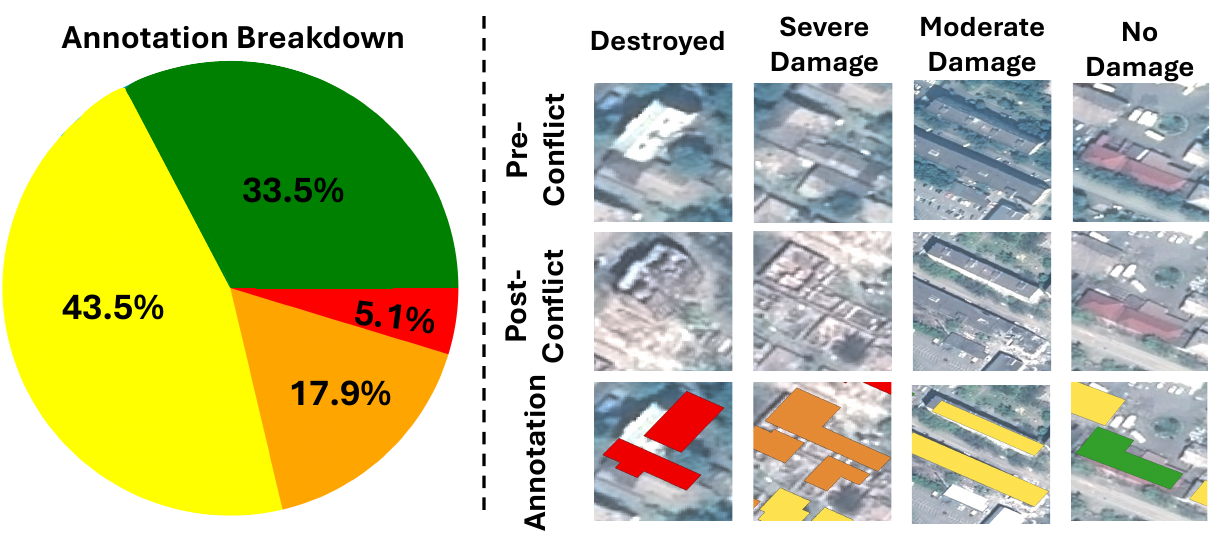}
  \vspace{-0.35cm}
  \caption{Annotation breakdown and examples.}
  \label{fig:annotations}
  \vspace{-0.6cm}
\end{figure}
\rev{The two images present a size of 8361$\times$13641 pixels with 6310 annotated buildings, of which 2405 buildings were classified using UNITAR information and the remaining 3905 by domain experts. Fig.~\ref{fig:annotations} shows the breakdown of the different annotations along with some annotation examples.}
%
%
\subsection{Network Architectures}
\begin{figure*}[t]
  \centering
  \includegraphics[width=0.9\textwidth]{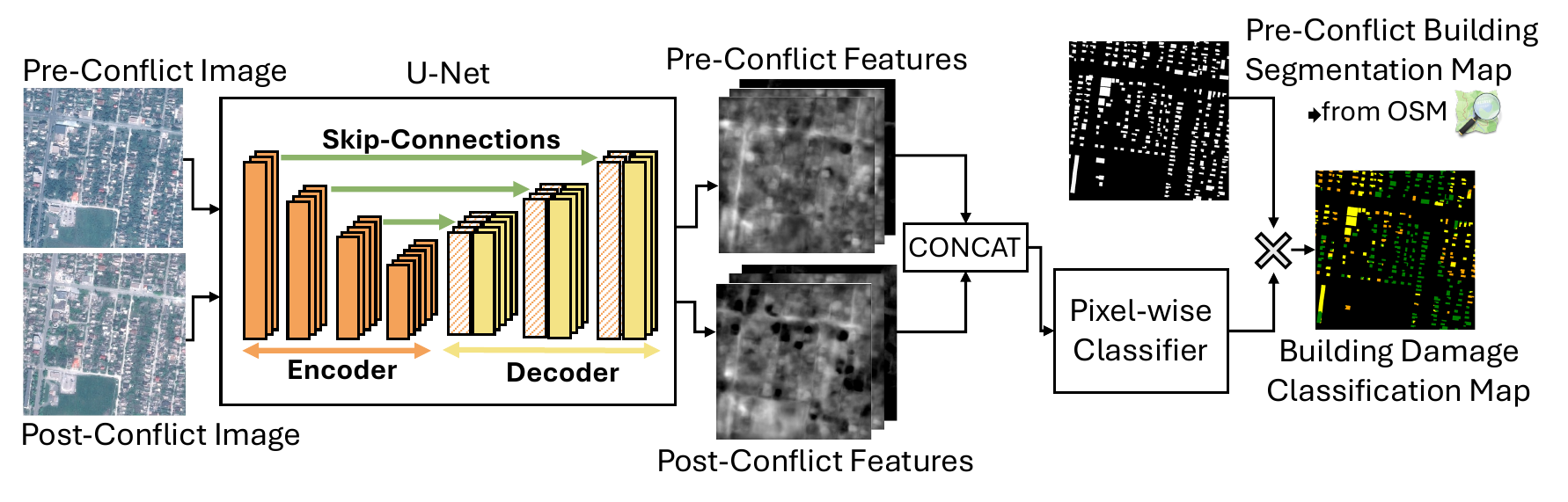}
  \vspace{-0.6cm}
  \caption{High-level overview of the proposed architecture based on the U-Net scheme.}
  \label{fig:arch}
\end{figure*}
Fig.~\ref{fig:arch} summarizes the proposed DL architecture.
We follow the scheme of~\cite{wheeler2020deep, shen2021bdanet, chen2022dual} in which pre- and post-disaster images are fed to a semantic segmentation model. The model builds a damage map with the same size as the two inputs, where each pixel is annotated with a label assessing the degree of the damage.
Typically, in the SotA~\cite{wheeler2020deep} the task of \textit{identifying buildings} and \textit{assessing their damage} is demanded to two separately trained networks. The first one performs a binary classification, associating each pixel of the pre-disaster image to a ``building/no-building'' label. The second DNN then performs the pixel-wise damage classification using paired pre and post images. Lastly, the two segmentation maps are point-wise multiplied to obtain the final building damage map.
In our case, however, given the availability of the buildings' footprint in the area of interest from the OSM database, \textit{we do not need the first network}, and we can generate the buildings' footprint directly leveraging the information provided by OSM.
\subsubsection{U-Net Architecture}~\label{subsec:u_net}
As depicted in Fig.~\ref{fig:arch}, we use segmentation models based on the well-known U-Net~\cite{unet_original} architecture. 
\rev{Initially devised for biomedical image segmentation, U-Net has been later applied to geospatial tasks such as land cover classification, urban mapping, change detection, and environmental monitoring~\cite{unet++,wheeler2020deep, shen2021bdanet}.}
In our case, U-Net is used to extract a set of feature maps of the same dimension from the pre and post images. The two sets are then concatenated across the channel dimension and fed to a pointwwise convolution (i.e., with kernel size equal to $1 \times 1$) with a number of output channels equal to the number of considered damage classes.
%
%

The U-Net DNN consists of three main parts: the Contracting Path (Encoder), the Expansive Path (Decoder), and the Skip Connections that bridge the two paths.
The \textit{Encoder} consists of a series of convolutional layers, each followed by an activation function and an optional pooling layer. Pooling (or strided convolutions) progressively reduce the spatial dimensions of the feature maps, allowing the network to capture features at different scales and reducing the computational load for subsequent layers. In this paper, we consider four different encoder implementations which are detailed in Sec.~\ref{subsec:backbones}.
The \textit{Decoder} performs a progressive upsampling of the feature maps obtained from the contracting path to reconstruct the spatial dimensions of the original input image. In our work, we keep the Decoder architecture constant, while varying the Encoder. We always consider an up-sampling factor of 2 with the nearest neighbor strategy, and each up-sampled feature map is fed to a convolutional layer followed by a ReLU activation function.
\looseness=-1
\textit{Skip Connections} are added between corresponding layers in the contracting and expansive paths, allowing the network to fuse information from different scales, preserving both local fine-grained details and high-level global context. Specifically, the feature maps from the contracting path are \textit{concatenated} with the upsampled feature maps in the expansive path. 
%
\subsubsection{Encoder Blocks}~\label{subsec:backbones}
\begin{figure}[t]
  \centering
  \includegraphics[width=0.8\columnwidth]{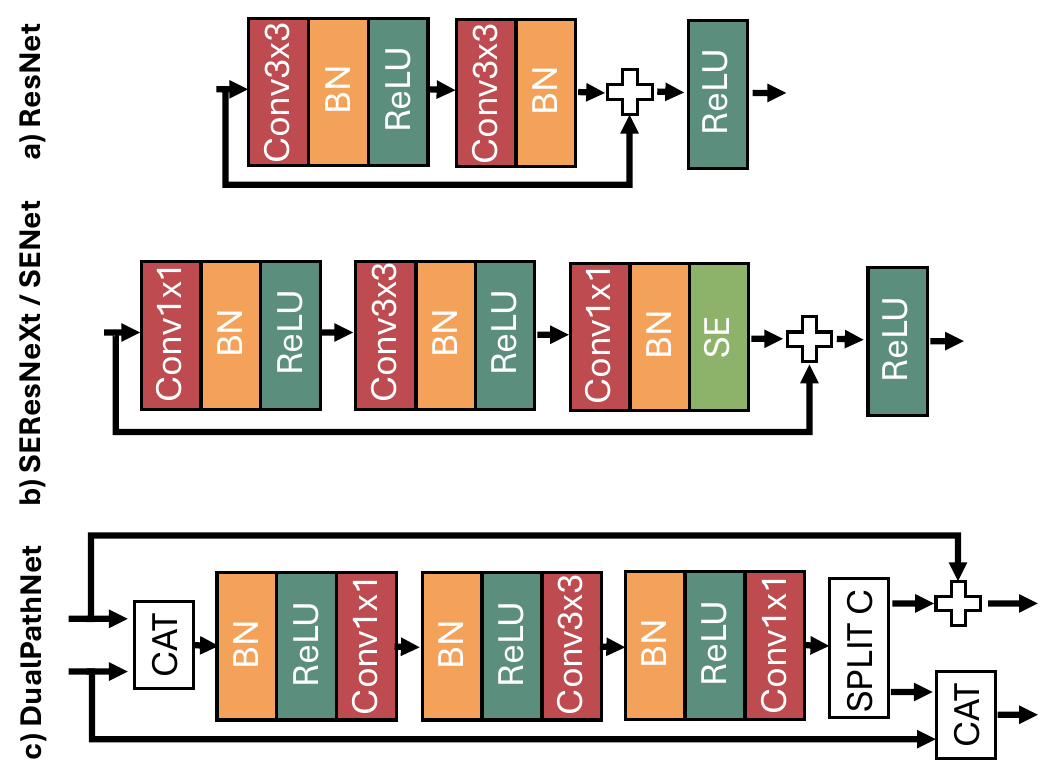}
  \vspace{-0.2cm}
  \caption{Basic Convolutional Blocks employed for the U-Net Encoder.}
  \label{fig:conv_blocks}
  \vspace{-0.6cm}
\end{figure}
In this work, we consider and compare four different convolutional blocks, shown in Fig.~\ref{fig:conv_blocks}, to build the U-Net encoder. Such networks are taken from the open-source implementation~\cite{xview_winner} winner of the xView2 challenge, thus representing the SotA for natural disaster damage detection.
The encoder is built by repeating each block five times, with different hyper-parameters. Moreover, a pooling layer is always added between the first and second block, while for the other blocks, spatial sub-sampling is achieved through strided convolution.

The simplest Encoder block is based on the ResNet scheme, depicted in Fig.~\ref{fig:conv_blocks}a. It includes two convolutional layers with 3$\times$3 kernel size, where the first is followed by Batch Normalization (BN) and ReLU activation and the second by BN only. The output of this stack is summed to the input through a residual connection and then passed through another ReLU. 
%
When the number of input and output features of a block differ, a pointwise convolution with BN is added on the residual path to adapt the output shape, making the elementwise sum possible.

A more complex scheme is the one employed by SEResNeXt and SENet encoders, depicted in Fig.~\ref{fig:conv_blocks}b. The two present the same architecture and differ only in the values of some hyper-parameters. The architecture is based on a bottleneck structure with a stack of three convolutional layers followed by BN and ReLU. The first convolution is pointwise and reduces the number of features, the second has a 3$\times$3 kernel which keeps the number of features constant in the case of SEResNeXt while in the case of SENet, it increases it by four times. The last convolution is again pointwise and it enlarges the features number by four times for SEResNeXt while keeping it constant for SENet. Moreover, the output of the last convolution is processed by a Squeeze-and-Excite (SE) module which adaptively recalibrates channel-wise feature importance~\cite{SE}. Finally, the output is again summed to the input of the block through a residual connection. As in the case of the ResNet block, if needed an additional pointwise convolution is added on the residual path.

The last option is the so-called DualPathNet (DPN) block, shown in Fig.~\ref{fig:conv_blocks}c. It takes two inputs which are concatenated along the channel dimension and fed to a similar bottleneck structure as the one of SEResNeXt and SENet. The output of this structure is split along the channel dimension in two parts, one is summed to the first input while the second one is concatenated with the second input. These two outputs then become the inputs of the subsequent block.
%
%
\subsection{Training scheme}~\label{subsec:training}
We partition the available satellite images into training and a test sub-images, ensuring no intersection between the two, to create a fair test environment for our evaluation. In particular, we adopt a Leave-One-Quarter-Out (LOQO) cross-validation approach, by dividing the image spatially in 4 quarters, iteratively keeping one of them as test set, and using the other 3 for training, in each fold. Namely, we generate the training dataset by creating partially overlapping image patches of dimension 1024$\times$1024, with a fixed stride. The partial overlap increases the number of training samples, while preserving data diversity since the same buildings are found at different (x, y) coordinates in different images.
On average, each training fold consists of approximately 10k image patches. The number varies slightly for each fold due to the non-rectangular area covered by the global image.

To deal with this relatively small dataset size, we implement an augmentation pipeline to create new plausible instances, with both geometric and photometric augmentations.
%
%
\rev{Each input image can be flipped either vertically, horizontally, or both. It can also be translated (with a random shift factor $\gamma_1 \in [-0.0625, 0.0625]$), scaled (with a random factor $\gamma_2 \in [0.9, 1.1] $) and rotated by an angle $\gamma_3 \in [-45^{\circ}, 45^{\circ}]$. Both the flipping and the other affine transformations are applied with a probability of 50\%. Next, three blocks of photometric augmentations are applied, each with a probability of 50\%. Within each block, only one transformation is applied, sampled uniformly. The first block applies either: i) a random shift in $[-20,20]$ to the values of each RGB channel; ii) a conversion to grayscale; iii) a sepia filter. The second block modifies either: i) the brightness and contrast of the image by a random factor in the range $[-0.2, 0.2$], or ii) applies a gamma correction, with a random factor drawn from the range $[80, 120]$. The third and last block comprises transformations that impact the image quality, namely: i) blurring with a randomly sized kernel (with the maximum being fixed at $[3, 7]$); ii) downscaling the image by a factor of 0.25 and sequentially upscaling it; iii) applying grid distortion, which divides the image patch into a grid and randomly modifies the intersection points, producing localized perturbations. The distortion factor lays in the range $[-0.3, 0.3]$.}
%

Due to slight differences in the off-nadir angle and in the sun-elevation angle, the images captured before and after a disruptive event may exhibit distinct looks that are not associated to the actual changes of interest (e.g., distortions in the appearance of tall buildings and wide, dark shadows alongside objects). To mitigate the impact of the unavoidable effects of the orbit of the satellites and of the time of day on the captured images, \rev{we also consider applying a dilation filter with a squared kernel of 3$\times$3 pixels, over the building segmentation map to thicken the buildings' footprints.}

As training objective, we employ a cross-entropy loss computed only on the annotated pixels, so that buildings for which no label mask is available does not contribute to the loss.
%

\section{Experimental Results}~\label{sec:results}
%
%
\begin{table*}[t]
\centering
\caption{DNN F1 score results over the Mariupol area. The ZS suffix refers to the models pre-trained only on the xBD dataset.}
\vspace{-0.2cm}
\label{tab:network_perf}
\resizebox{0.9\textwidth}{!}{%
\begin{tabular}{l|c|c:ccc|c|c:cc}
\hline
\textbf{Network} & \revtwo{$\mathbf{BAS_3}$} & $\mathbf{F1_3}$ & \textbf{No Damage} & \textbf{Moderate Damage} & \textbf{Severe + Destroyed} &  \revtwo{$\mathbf{BAS_2}$} & $\mathbf{F1_2}$ & \textbf{No Damage} & \textbf{Damage} \\ \hline
ResNet-ZS & \revtwo{74\%} & 38\% & 42\% & 31\% & 44\% & \revtwo{81\%} & 56\% & 42\% & 85\% \\
ResNet & \revtwo{\textbf{79\%}} & \textbf{59\%} & 54\% & 72\% & 54\% & \revtwo{85\%} & 66\% & 54\% & 85\% \\ \hline
SEResNeXt-ZS & \revtwo{77\%} & 47 \% & 51 \% & 42 \% & 49 \% & \revtwo{84\%} & 60 \% & 51 \% & 74 \% \\
SEResNeXt & \revtwo{76\%} & 46 \% & \textbf{57} \% & 70 \% & 30 \% & \revtwo{\textbf{86\%}} & \textbf{69} \% & \textbf{57} \% & 86 \% \\ \hline
SENet-ZS & \revtwo{74\%} & 43 \% & 42 \% & 42 \% & 46 \% & \revtwo{81\%} & 57 \% & 42 \% & 86 \% \\
SENet & \revtwo{72\%} & 42 \% & 40 \% & 67 \% & 32 \% & \revtwo{80\%} & 54 \% & 40 \% & 84 \% \\ \hline
DualPathNet-ZS & \revtwo{74\%} & 39 \% & 43 \% & 32 \% & 43 \% & \revtwo{81\%} &  56 \% & 43 \% & 79 \% \\
DualPathNet & \revtwo{78\%} & 55 \% & 45 \% & 72 \% & \textbf{56} \% & \revtwo{82\%} & 59 \% & 45 \% & 87 \% \\ \hline
Ensemble-ZS & \revtwo{77\%} & 44 \% & 53 \% & 35 \% & 48 \% & \revtwo{84\%} & 65 \% & 53 \% & 84 \% \\
Ensemble & \revtwo{78\%} & 58 \% & 55 \% & \textbf{73} \% & 50 \% & \revtwo{85\%} &  67 \% & 55 \% & \textbf{88} \% \\

\end{tabular}
}
\vspace{-0.4cm}
\end{table*}
\begin{table*}
\centering
\caption{Comparison of the ResNet training performance on our dataset with and without xBD pre-training.}
\vspace{-0.2cm}
\label{tab:pretrain_eff}
\resizebox{0.8\textwidth}{!}{%
\begin{tabular}{c|c:ccc|c:cc}
\hline
\textbf{Pretraining} & $\mathbf{F1_3}$ & \textbf{No Damage} & \textbf{Moderate Damage} & \textbf{Severe + Destroyed} & $\mathbf{F1_2}$ & \textbf{No Damage} & \textbf{Damage} \\ \hline
\ding{55} & 56\% & 52\% & \textbf{73\%} & 49\% & 65\% & 52\% & \textbf{87\%} \\
\ding{51} & \textbf{59\%} & \textbf{54\%} & 72\% & \textbf{54\%} & \textbf{66\%} & \textbf{54\%} & 85\% \\
\end{tabular}
}
\vspace{-0.4cm}
\end{table*}
\begin{table}[t]
\centering
\caption{Ablation study of the effect of Augmentations and Dilation filter for the ResNet architecture.}
\vspace{-0.2cm}
\label{tab:aug_dil_abl}
\resizebox{0.9\columnwidth}{!}{%
\begin{tabular}{c|c|c|c}
\hline
 \textbf{Metric} & $\mathbf{Baseline}$ & \textbf{+ Augmentation} & \textbf{+ Dilation}  \\ \hline
\textbf{$F1_3$} & 53\% & 58\% & \textbf{59\%} \\ \hline
\textbf{$F1_2$} & 61\% & 65\% & \textbf{66\%} \\

\end{tabular}
}
\vspace{-0.4cm}
\end{table}
\vspace{-1cm}
\subsection{Setup}
All our code is written in Python v3.9, using the PyTorch v2.3 library. Augmentations are implemented using albumentations v1.4 and geospatial images are managed using rasterio v1.3.9. All experiments are conducted on a workstation with a 32-core Intel Xeon w5-3435X and four NVIDIA A5000 GPUs.

We evaluate our DNN models on the dataset described in Sec.~\ref{subsec:dataset}. We consider both a \textit{Zero-Shot} (ZS)~\cite{zero_shot_survey} scenario with models pre-trained on xBD and tested on the Ukrainian data, and fine-tuning/training-from-scratch scenarios, where we employ the LOQO cross-validation scheme described in Sec.~\ref{subsec:training}.
For pre-trained models, we directly use the open-source weights published by the xView2 challenge winning entry~\cite{xview_winner}. For this reason, in all our experiments we set the input-images size to be the one expected by such models, i.e., 1024$\times$1024 pixels.
We consider an overlap of 960 pixels between images.
All models are trained for 30 epochs with a batch size of 64. 
%
%
The optimizer is AdamW with a weight-decay set to 5e-6 and an initial learning rate of 5e-5.

\rev{Following the SotA~\cite{xBD, shen2021bdanet, wheeler2020deep, chen2022dual} we employ the F1 score  as the main performance metric.}
%
%
For each class $C_i$, the class-wise $F1_{C_i}$ score is defined as
\begin{equation}
    F1_{C_i} = \frac{2TP_{C_i}}{2TP_{C_i} + FP_{C_i} + FN_{C_i}}   
\end{equation}
\vspace{-0.05cm}
with $TP_{C_i}$, $FP_{C_i}$, and $FN_{C_i}$ respectively the number of True Positive, False Positive, and False Negatives for the $i$-th class.
\revtwo{Moreover, we also consider the Balanced Accuracy Score (BAS), defined as the average recall (i.e., $TP/(TP + FN)$) obtained on each class.}
The F1 score reported in all our results is the average over the four quarters of the image. For trained models, we compute the predictions relative to each quarter using a model trained on the other 3.
\subsection{Results Analysis}
Table~\ref{tab:network_perf} summarizes the results obtained when testing the four considered U-Net-based models on our dataset. Moreover, we consider an ensembling approach where the final prediction is built as the average prediction of the four models.
For each model, we compare its ZS performance with the results obtained after training with the LOQO scheme.
Although the dataset has been annotated with four classes of damage, here we consider two distinct problems with 3 and 2 classes respectively. 
The 3-class version merges ``Severe-Damage'' and ``Destroyed'' annotations in the same class, and aims at distinguishing them from ``No-Damage'' and ``Moderate-Damage''.
The 2-class problem, instead, is aimed at identifying damaged vs undamaged buildings, merging all classes except ``No-Damage'' into one.
\looseness=-1
\rev{Reducing the number of classes simplifies the DNN's task and aligns with the needs of humanitarian organizations, which prioritize identifying heavily damaged areas for rapid intervention. It also mitigates the impact of noisy annotations, as distinguishing between destroyed and severely damaged buildings can be arbitrary even for expert human labelers.}

\rev{Different insights can be drawn from these results. First, the ZS networks perform better than random (i.e., F1 score $>$33\% and $>$50\% for 3- and 2-class problems respectively), demonstrating a certain degree of transferability. However, the networks trained with the Ukrainian data outperform the ZS networks in the majority of cases, underscoring the importance and effectiveness of retraining.}
On the more challenging 3-class problem, the trained ResNet improves the F1 score by 21\% over the ZS version, while on the easier 2-class scenario the same network improves the F1 score by 10\%.
The only architecture where training never improves the ZS results is the SENet. The reason for this behavior can be found in the fact that SENet is the largest network, thus more prone to overfitting especially when the volume of data is not so large, as in our case.
Coherently, the smallest architecture, i.e., the ResNet achieves the best results on the 3-class problem with an F1 score of 59\%. Instead, the SEResNeXt architecture achieves the best F1 score in the two-class scenario with 69\%.

Interestingly, we can notice how the F1 scores on the No-Damage class are on par between the ZS and trained networks while the ZS networks perform poorly on the Moderate-Damage class. \rev{This is expected, as identifying No-Damage does not depend on the type of scenario considered (natural disaster or war), allowing ZS networks to perform well. In contrast, war damage differs significantly from the one caused by natural disasters, leading to poorer performance of the ZS networks in identifying different degrees of damage.}

\revtwo{Finally, it is important to point out that the ground-truth annotations are performed at the single building level, while our models classify pixel by pixel. Therefore, aggregating the predictions over an entire building (e.g., using some representative statistic) could further improve performance. We plan to address this issue in the future works.}
\subsection{Ablation Studies}
In this section, we assess the importance of the different parts of the proposed pipeline. Namely, we first compare the results obtained fine-tuning DNNs originally pre-trained on xBD with training from scratch. Then, we perform an ablation study to investigate the effect of data augmentations and of the dilation filter discussed in Sec.~\ref{subsec:training}. The whole analysis is performed on the ResNet architecture, i.e., the best-performing architecture on the 3-class problem.

Table~\ref{tab:pretrain_eff} summarizes the effect of the pre-training on the xBD dataset (i.e., transfer learning). With respect to training from scratch, pretraining helps in boosting the performance for both versions of the problem. In particular, with 3 classes, the F1 score increases from 56\% to 59\% with a net improvement of 3\%. The improvement on the 2-class version is still present but more limited (+1\%).

Table~\ref{tab:aug_dil_abl} compares a baseline trained with the scheme of Sec.~\ref{subsec:training} with and without the proposed augmentation pipeline and the dilation filter. The most important contribution to the F1 score is represented by our proposed custom augmentations, which enhances performance by 6\% and 4\% respectively for 3- and 2-class scenarios. The dilation filter helps to further improve the F1 score by 1\% in both cases.
%

\section{Conclusions}~\label{sec:conclusions}
DNN-enabled automatic classification of VHR geospatial imagery represents a key technology to improve HADR, especially in war-related scenarios which are under-explored. 
\rev{This work employs an annotated dataset for building damage assessment, with images acquired over the Mariupol city, to train and assess the performance of a collection of SotA CNNs originally proposed for natural disasters.}
Future works will include the study of novel architectures such as Transformers and the collection and the annotation of more data.

\tiny
\bibliographystyle{IEEEtran}
\bibliography{bstctl,references}

\end{document}